\documentclass[11pt]{article}

\usepackage[final]{acl}
\usepackage{times}
\usepackage{latexsym}
\usepackage[T1]{fontenc}

\usepackage[utf8]{inputenc}

\usepackage{microtype}

\usepackage{inconsolata}

\usepackage{graphicx}

\usepackage{booktabs}
\usepackage{makecell}
\usepackage{threeparttable}
\usepackage{pifont}
\usepackage{tabularx}
\usepackage{amsmath}
\usepackage{subcaption}
\usepackage{amssymb}

\usepackage[most]{tcolorbox}
\tcbuselibrary{skins}
\usepackage{textgreek}

\usepackage[table]{xcolor}

\usepackage{tipa}

\newcommand{\cmark}{\ding{51}}%

 \newcommand{\rc}[1]{\cellcolor{blue!#1}}

\usepackage[dvipsnames]{xcolor}

\title{Speculative Decoding and the Curse of Multilinguality}

\author{Nirajan Paudel* \and Michael Ginn* \and Luc De Nardi \and Alexis Palmer \\
       University of Colorado \\ *Equal contribution  }

\begin{document}
\maketitle
\begin{abstract}
Speculative decoding \citep{leviathan2023fastinferencetransformersspeculative, chen2023acceleratinglargelanguagemodel} is a popular technique for large language model (LLM) inference, enabling faster generation by drafting token sequences
with a smaller \textit{draft model}. However, the effectiveness of speculative decoding has mainly been studied for English. Motivated by the \textit{curse of multilinguality}, we hypothesize that speculative decoding is far less effective for low-resource languages due to the limited multilingual capacities of smaller models. We test nine low-resource languages and four high-resource languages under a standard speculative decoding setup and find moderate evidence for our hypothesis. 

Next, we try to improve the multilingual capabilities of the smaller draft model via distillation from the larger model. We find, though, that distillation generalizes poorly across tasks in the same language, and we argue that assembling a task-agnostic, fully representative dataset is infeasible for low-resource languages. Finally, we propose weaker n-gram models as draft models; these provide moderate speed-ups due to their minuscule inference cost.


\end{abstract}

\section{Introduction}
\subsection{Speculative Decoding...}

\begin{figure}[tbh!]
    \centering
    \includegraphics[width=\linewidth]{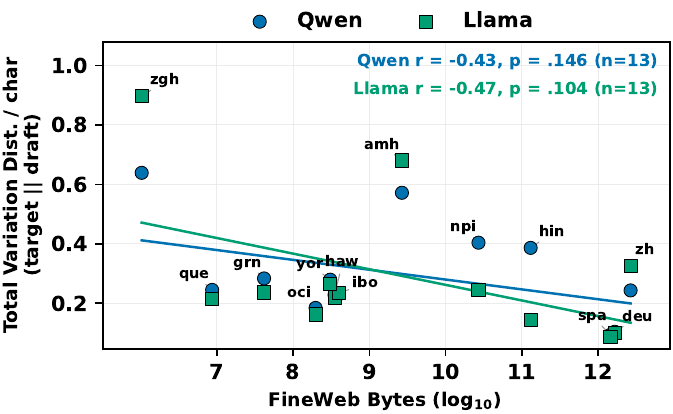}
    \caption{Relationship between language resourcedness (estimated via bytes in FineWeb2) and total variation distance (LK divergence) between a larger and smaller model, computed on monolingual text in each language. \textbf{Higher-resource languages tend to have a lower divergence between models.}}
    \label{fig:lk_divergence_vs_resourcedness}
\end{figure}

Autoregressive decoding with a language model requires $K$ serial forward passes to generate a sequence of $K$ tokens. \textit{Speculative decoding} is a popular technique to accelerate inference by first generating a short sequence of tokens from a smaller \textit{draft model} and then verifying the drafted tokens in parallel \citep{leviathan2023fastinferencetransformersspeculative, chen2023acceleratinglargelanguagemodel}, which can drastically reduce the total number of forward passes with the larger target model (also known as the \textit{verifier}). 

The critical metric for speculative decoding efficacy is the \textbf{expected speed-up factor}, computed from \textbf{acceptance rate} ($\alpha$) and the draft and verifier models' \textbf{inference costs} (time). Acceptance rate is the probability of accepting a token from the draft model; a low acceptance rate results in a low speed-up factor, or even a slowdown. In order to achieve a high acceptance rate, the next-token distributions of the draft and verifier model must be very similar; i.e. they should have a low \textbf{divergence}.

\subsection{...And The Curse of Multilinguality}
Modern large language models are effective across a variety of languages, largely due to the \textit{crosslingual transfer} of capabilities from higher-resource to lower-resource languages \citep{wu-dredze-2019-beto,wu-dredze-2020-languages}. However, training on data from many languages does not guarantee positive transfer. In some cases, training on highly multilingual datasets can actually impede crosslingual transfer, a phenomenon known as the \textbf{curse of multilinguality} \citep{conneau-etal-2020-unsupervised}. This issue arises because although models share many parameters across languages, there are an irreducible number of language-specific parameters, so adding more languages can exhaust the model's representational capacity \citep{wang-etal-2020-negative}. Naturally, then, the curse disproportionately affects smaller models with less overall capacity \citep{chang-etal-2024-multilinguality}, while large-scale LLMs may show little evidence of degradation \citep{foroutan2025revisitingmultilingualdatamixtures}. 

The curse also affects languages differently based on resourcedness \citep{chang-etal-2024-multilinguality}. For high-resource languages, smaller models effectively learn lower-fidelity language modeling distributions which are \textbf{less accurate but still similar to the distributions of larger models}. Meanwhile, in the low-resource regime, smaller models lose out on gains from crosslingual transfer entirely, diverging greatly from larger models that learn robust crosslingual representations. We can quantify this difference with divergence metrics (such as KL divergence) over the next-token prediction distribution, and we pose the following hypothesis:

\paragraph{Hypothesis 1:} \textit{Small models diverge more from large models for lower-resource languages.}

\paragraph{} This is a critical issue for speculative decoding, 
since acceptance rates directly depend on the divergence between a smaller draft model and larger verifier model. Thus, we form a second hypothesis:

\paragraph{Hypothesis 2:} \textit{Acceptance rates are far lower for lower-resource languages, resulting in unfavorable speed-ups.} 




\paragraph{} 
Poor speed-ups mean non-English users may suffer far slower generation, yet another instance of disparities in language technologies \citep{blasi-etal-2022-systematic}. The issue is exacerbated by tokenization biases, where less common languages often require far more tokens per output \citep{NEURIPS2023_74bb24dc,mehta2026defragmenting}. 

\subsection{Contributions}
We present a \textbf{scientific investigation of the relationship between language resourcedness, divergence, and speculative decoding}. We provide empirical support for both hypotheses, showing high divergence and low acceptance rates when generating outputs in low-resource languages (\autoref{fig:lk_divergence_vs_resourcedness} and \S \ref{sec:results}). We note the difficulty of creating an evaluation dataset that fully covers the distribution of real-world usage in a language (\S \ref{sec:datasets}), so we select two representative tasks for evaluation: translation and story generation. We benchmark speculative decoding metrics using pairs of multilingual verifier and draft models from the Qwen 3.5 and Llama 3.2 families on \textbf{nine low-resource languages and four high-resource languages}. We make the following contributions:
\begin{itemize}
    \item We provide evidence that draft/verifier divergence, and therefore speculative decoding effectiveness, is negatively correlated with language resourcedness (\S \ref{sec:results}).
    \item We evaluate speculative decoding after \textbf{distillation} from the verifier model to the draft model, which tends to help on prompts similar to the training dataset, but generalizes poorly to new settings (\S \ref{sec:distillation}). 
    \item We propose the use of n-gram draft models, which exhibit far lower acceptance rates but near-zero inference costs (\S \ref{sec:ngrams}).
\end{itemize}
\noindent We release our code on GitHub.\footnote{\url{https://github.com/michaelpginn/spec-decoding}}

\begin{table*}[tb!]
    \small
    \centering
        \begin{tabular}{l c c c c | c c}
            \toprule
            \textbf{Language} & \multicolumn{2}{c}{\textbf{Mono (\# toks)} } & \multicolumn{2}{c|}{\textbf{Parallel (\# sents)}} & \multicolumn{2}{c}{\textbf{Estimated Resourcedness}} \\
            ~[iso code] & Train & Test & Train & Test & \# words & \# bytes \\
            \midrule
            Tamazight [zgh] & 4.8M   & 336.6k & 4800 & 400 & \rc{8} 83k   & \rc{8} 1.0 MB \\
            Quechua [que]   & 510.2k & 131.8k & 4800 & 400 & \rc{16} 1.3M & \rc{16} 8.7 MB \\
            Guarani [grn]   & 319.3k & 79.2k  & 788  & 198 & \rc{21} 6.6M & \rc{21} 42 MB \\
            Occitan [oci]   & 46.0k  & 11.5k  & 3631 & 400 & \rc{27} 36M  & \rc{27} 200 MB \\
            Yoruba [yor]    & 357.8k & 91.8k  & 4800 & 400 & \rc{28} 59M  & \rc{28} 310 MB \\
            Hawaiian [haw]  & 82.1k  & 20.5k  & 96   & 25  & \rc{29} 75M  & \rc{29} 350 MB \\
            Igbo [ibo]      & 677.5k & 164.0k & 1398 & 350 & \rc{29} 76M  & \rc{29} 400 MB \\
            Amharic [amh]   & 111.7k & 36.8k  & 4800 & 400 & \rc{36} 240M & \rc{36} 2.7 GB \\
            Nepali [npi]    & 13.0M  & 3.2M   & 3133 & 400 & \rc{44} 1.6B & \rc{44} 27 GB \\
            \midrule
            Hindi [hin]     & 24.6M  & 6.1M   & 4800 & 400 & \rc{49} 11B  & \rc{49} 130 GB \\
            Spanish [spa]   & 12.7M  & 3.4M   & 4800 & 400 & \rc{58} 262B & \rc{58} 1.4 TB \\
            German [deu]    & 12.2M  & 3.0M   & 4800 & 400 & \rc{58} 262B & \rc{58} 1.7 TB \\
            Chinese [zh]    & 14.2M  & 3.4M   & 4800 & 400 & \rc{60} 544B & \rc{60} 2.7 TB \\
            \bottomrule
        \end{tabular}
    \caption{Per language, number of tokens of monolingual text and number of parallel sentences in our datasets sorted by resourcedness. Sources are described in \autoref{tab:mono_source_counts} and \autoref{tab:par_source_counts}. We provide the number of words/bytes in Fineweb2 as a rough estimate of language resourcedness;  cell shading is proportional to $\log$ bytes. The top block contains the nine low-resource languages and the bottom block the four high-resource languages.}
    \label{tab:combined}
\end{table*}


\section{Background and Related Work}
\citet{leviathan2023fastinferencetransformersspeculative, chen2023acceleratinglargelanguagemodel, xia-etal-2023-speculative} proposed speculative decoding as a technique for accelerating generation. Specifically, the draft model autoregressively generates $\gamma$ new tokens with conditional probabilities $q(x_1) \dots q(x_\gamma)$, and the verifier model runs a single forward pass over the full drafted sequence, assigning distinct probabilities $p(x_1) \dots p(x_\gamma)$. For each token $i$, if $q_i\leq p_i$, the token is accepted. If $q_i > p_i$, we randomly choose  between: a) accepting the token with probability $\frac{p_i}{q_i}$; or b) rejecting the token and all subsequent draft tokens and then sampling a new token from the adjusted distribution $q'=p(x)-q(x)$. This results in a sampling distribution equivalent to the verifier model's distribution.

To improve the probability of accepting a draft token, a popular technique is performing \textit{knowledge distillation} from the verifier model to the draft model \citep{zhou2024distillspec}, where the draft model is finetuned to minimize its KL divergence with the teacher's distribution. \citet{yi-etal-2024-towards} apply this technique to a multilingual setting, distilling models for translation from another language into English, but they do not study the more difficult case of generating text in the target language.\footnote{Except for a brief evaluation in the appendix.}

Our work is most similar to \citet{sandler2025disparateimpactsspeculativedecoding}, which measures acceptance rate disparities across tasks and languages. They propose methods to mitigate these issues by balancing the training dataset or scaling per-task gradients during distillation. However, their approach requires a representative dataset for each of the desired tasks, which is infeasible to acquire for many languages. Instead, we focus on testing whether methods can \textbf{generalize to new tasks without task-specific training data}.

\section{Methodology}

\subsection{Models}
Our main results (\S\ref{sec:results}) are collected with models from the Qwen 3.5 family \citep{qwen35blog} and the Llama 3.2 family \cite{grattafiori2024llama3herdmodels}. For Qwen, we use the 9b parameter model as the verifier and the 0.8b parameter model as the drafter. For Llama, we use the 3b parameter model as the verifier and the 1b parameter model as the drafter. We also run a subset of experiments with the Qwen 2b and 4b parameter draft models in \S \ref{sec:acc_rates_ap}.

\subsection{Datasets}
\label{sec:datasets}
Benchmarking speculative decoding for a given language requires a representative test dataset. However, assembling a dataset that fully covers the domain of real-world usage is virtually impossible, particularly for low-resource languages, and prior work typically uses task-specific datasets for tasks like question answering \citep{sandler2025disparateimpactsspeculativedecoding}. We use datasets for nine low-resource languages and four high-resource languages, covering a range of resourcedness, typological properties, and geographic regions. We assemble three types of dataset per language: 1) monolingual, domain-agnostic data; 2) parallel sentences (with English) for translation; and 3) prompts for generating open-ended stories. Of these, the test splits of \textbf{2 and 3 are used to test speculative decoding efficiency}, while the train splits are reserved for draft model training and distillation.

Both the monolingual and parallel sentence corpora are collected from a variety of sources described in \autoref{sec:sources}. Dataset sizes are summarized in \autoref{tab:combined}; the monolingual corpora range from 46k tokens to 25M tokens (under the Qwen tokenizer), while the parallel corpora are limited to a maximum of 5,200 examples per language.
For story generation, we create a dataset of 200 topics (e.g., \textit{wooden cottage}, \textit{wild pet}, \textit{young detective}, \textit{distant monastery}) by pairing concrete nouns from \citet{brysbaert2014concreteness}, filtered with WordNet \citep{miller1995wordnet} to people, animals, and places, with
adjectives from the Brown Corpus \citep{francis1979brown}, according to the similarity of their GloVe word embeddings \citep{pennington-etal-2014-glove}. 

We also estimate \textbf{language resourcedness} in order to study its relationship to divergence and acceptance rates. FineWeb2, an open-source pretraining dataset sourced from Common Crawl \citep{penedo2025fineweb2}, reports the number of words and bytes per language calculated using a language classifier (also in \autoref{tab:combined}). We use the number of bytes\footnote{Rather than the number of words, which is highly dependent on linguistic properties of each language.} as a rough proxy for language resourcedness. Of course, this dataset may not be fully representative of the pretraining data used for Qwen/Llama, but those datasets are not publicly available. 



\subsection{Tasks}
We test speculative decoding for two tasks: machine translation and story generation. For translation, we prompt the model to translate \textbf{from English into the target language}, since we care about decoding efficiency in languages other than English: 
\begin{tcolorbox}[colback=blue!5!white,colframe=blue!75!black]
\small
  Translate the following English text to \textbf{<language>}. Output only the translation, nothing else. \textbf{<English text>}
\end{tcolorbox}
\noindent 
The accuracy of the generated translation is not a primary concern for our experiments, but we do report these scores in \S\ref{sec:appendix_translation_metrics}.

For story generation, we prompt the model to generate one target language story per topic:
\begin{tcolorbox}[colback=blue!5!white,colframe=blue!75!black]
\small
  Write a short story in \textbf{<language>} about a(n) \textbf{<topic>}. Output only the story, nothing else.
\end{tcolorbox}
\noindent A few examples of generated stories appear in \autoref{fig:story-examples}. We do not compute any metrics over these stories, but we do verify that they are consistently generated in the correct language and do not get stuck in endless repetition loops.


\subsection{Evaluation}
\label{sec:evaluation}

To evaluate speculative decoding efficiency, we generate outputs from the verifier model using the prompts in the prior section, formatted with the model's chat template. We use sampling-based inference with a top-k of 50, top-p of 0.9, repetition penalty of 1.1 (16 token window), and a maximum 128 generated tokens. We use KV caching for both the draft and verifier models \citep{pope2022efficientlyscalingtransformerinference}. For each setting, we sweep the $\gamma$ value over $[2,4]$ and report the best result. During generation, we compute the following standard metrics:

\paragraph{Acceptance Rate ($\alpha$).}
Following \citet{leviathan2023fastinferencetransformersspeculative}, Definition~3.1, we measure the probability that a draft token is accepted by the target model, given all prior tokens have been confirmed. This probability is the complement of the expected divergence between the two model distributions:
\begin{equation}
    \alpha = 1 - E[D_{L,K}(p,q)]
\end{equation}
where 
\begin{equation}
\label{eq:lk}
    D_{L,K} = \sum_x\frac{|p(x)-q(x)|}{2}
\end{equation}
$D_{L,K}$ is a natural divergence between the teacher model ($p$) and the student model ($q$); it is in fact exactly the \textit{total variation distance}. Further, through Pinsker's inequality $D_{L,K}$
is upper-bounded by the more familiar KL divergence $D_{KL}$:\footnote{The similarity in notation is coincidental.}
\begin{equation}
\label{eq:bound}
    D_{L,K}\leq \sqrt{\frac{1}{2}D_{KL}}
\end{equation}
The proof is given in \S\ref{sec:kl_div}. This relationship has implications for distillation, discussed in \S \ref{sec:distillation}.

In practice, we estimate the acceptance rate for a single example $i$ with the Monte Carlo estimator $\hat{\alpha_i} = \frac{m_i}{d_i}$ where $m_i$ is the number of accepted tokens and $d_i$ is the number proposed, and we take the mean over examples.



\begin{figure}[b!]
    \centering
    \begin{subfigure}{\linewidth}
    \includegraphics[width=1\linewidth]{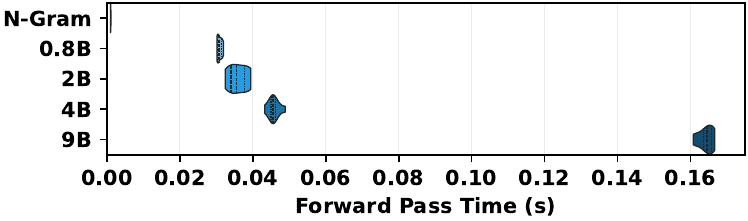}
    \end{subfigure}
    \begin{subfigure}{\linewidth}
    \includegraphics[width=1\linewidth]{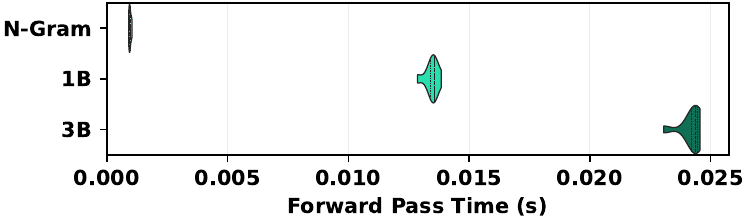}
    \end{subfigure}
    \caption{Distribution of inference costs (time for a single forward pass) for Qwen (top), Llama (bottom) and n-gram models. 
    \textbf{The small line in the upper left represents the n-gram time of 0.001s.}}
    \label{fig:fwd_pass}
\end{figure}

\paragraph{Inference Cost Ratio (c).}
We also compute the inference cost ratio, defined as the ratio between the time cost of a single forward pass for the draft model and a forward pass for the verifier model:
\begin{equation}
    c = t_{\text{draft}} / t_{\text{target}}
\end{equation}
We measure these costs on real hardware (H100 GPUs) using CUDA timing. We report the costs for all models in this study in \autoref{fig:fwd_pass}.

\paragraph{Speed-up Factor ($f$).}
Finally, we compute the speed-up factor, which quantifies the expected improvement under speculative decoding using the acceptance rate and inference cost ratio:
\begin{equation}
f = \frac{1 - \alpha^{\gamma+1}}{(1 - \alpha)(\gamma c + 1)}
\end{equation}
Speed-up values are reported as multiplicative factors (e.g., $1.2\times$).

\begin{figure}[tb!]
    \centering
    \includegraphics[width=\linewidth]{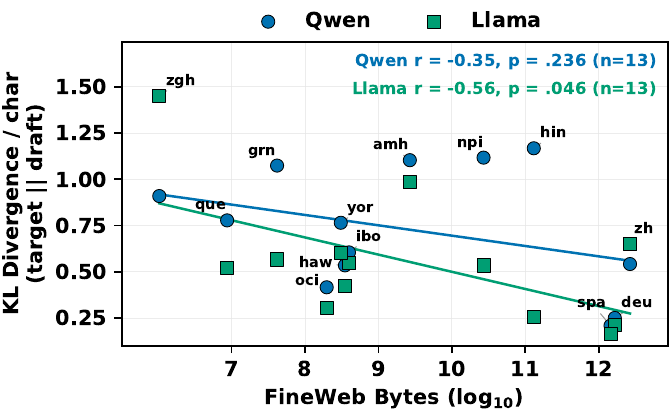}
    \caption{Relationship between language resourcedness (bytes in FineWeb2) and KL divergence between a larger and smaller model, computed on monolingual text in each language.}
    \label{fig:kl_divergence_vs_resourcedness}
\end{figure}
\begin{figure*}[t!]
    \centering
    \begin{subfigure}{0.49\textwidth}
        \includegraphics[width=1\linewidth]{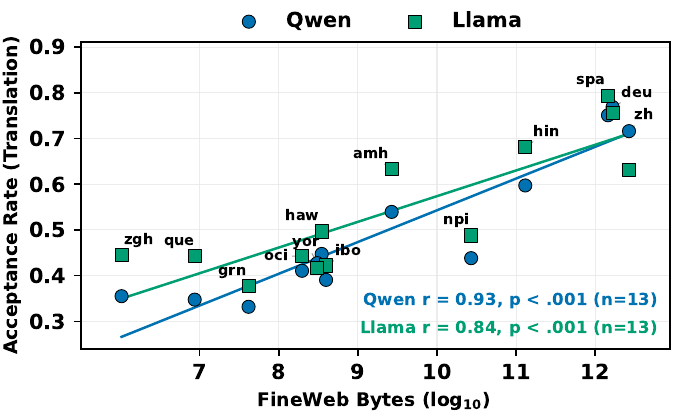}
    \end{subfigure}
    ~
    \begin{subfigure}{0.49\textwidth}
        \includegraphics[width=1\linewidth]{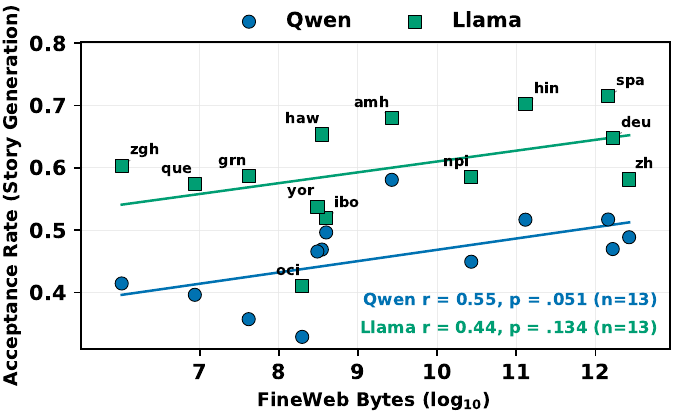}
    \end{subfigure}
    \caption{Relationship between language resourcedness (number of bytes in FineWeb2) and speculative decoding acceptance rates for \textbf{translation (left)} and \textbf{story generation (right)}.}
    \label{fig:baseline_acceptance}
\end{figure*}
\begin{figure*}[t!]
    \centering
    \begin{subfigure}{\linewidth}
    \includegraphics[width=1\linewidth]{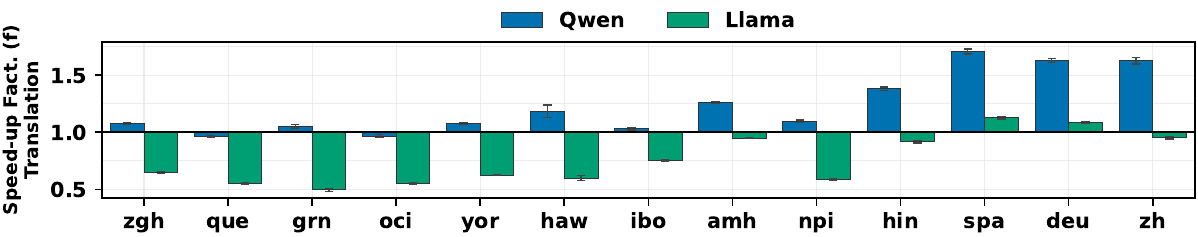}
    \end{subfigure}
    \begin{subfigure}{\linewidth}
    \includegraphics[width=1\linewidth]{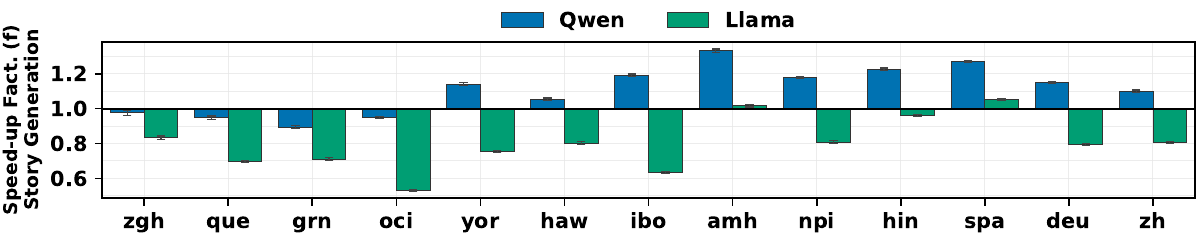}
    \end{subfigure}
    \caption{Speculative decoding speed-up factors for \textbf{translation (top)} and \textbf{story generation (bottom)}. The line at 1.0 indicates no speed-up over the baseline, and values below the line indicate a slow-down. Languages are ordered by resourcedness.}
    \label{fig:baseline_speedups}
\end{figure*}

\section{Results}
\label{sec:results}

\subsection{Domain-Agnostic Divergence}
\label{sec:results_divergence}
To test \textbf{Hypothesis 1}, we first measure the divergence between the next-token predictions of the verifier model and the draft model over the monolingual corpora, providing a coarse estimate of domain-agnostic behavior. Note that this is an off-policy evaluation, where all tokens come from a pre-existing document, rather than being generated by the model itself. We measure two divergences: 1) the standard KL divergence from the verifier to the draft model and 2) the LK divergence as defined in \autoref{eq:lk}; we normalize these values per-character to control for tokenization disparities. \autoref{fig:lk_divergence_vs_resourcedness} and \autoref{fig:kl_divergence_vs_resourcedness}  visualize the relationship between language resourcedness and LK/KL divergence, respectively. 
Although we use a very rough proxy for language resourcedness, the results provide evidence for Hypothesis 1,
with moderate negative correlations indicating that lower-resource languages tend to show higher divergences.

\subsection{Speculative Decoding Acceptance Rates}
\label{sec:acc_rates_ap}
To test \textbf{Hypothesis 2}, we compute acceptance rates under the speculative decoding setup described in \S\ref{sec:evaluation}. This is an on-policy evaluation, as acceptance rate is calculated over tokens generated by the verifier model. We report mean per-token acceptance rates, as a function of language resourcedness, in \autoref{fig:baseline_acceptance}. We make the following observations:

\paragraph{Model pairs with a smaller size difference have higher acceptance rates.} Qwen generally has lower acceptance rates than Llama on both tasks. This result is intuitive, as the Qwen models (9b and 0.8b) have a much greater size disparity than the Llama models (3b vs 1b).

\begin{figure}[tbh!]
    \centering
    \begin{subfigure}{\linewidth}
        \includegraphics[width=\linewidth]{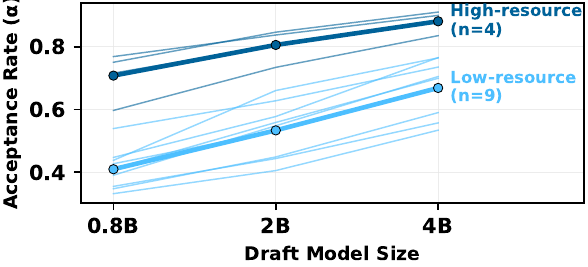}
    \end{subfigure}
    \begin{subfigure}{\linewidth}
        \includegraphics[width=\linewidth]{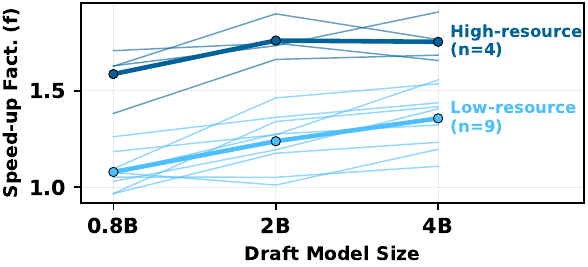}
    \end{subfigure}
    \caption{Acceptance rates (top) and speed-up factors (bottom) for \textbf{translation prompts} with \textbf{Qwen-based} draft models of various sizes.}
    \label{fig:draft_model_size_results}
\end{figure}

To control for differences across model families, we also test larger Qwen draft models and report acceptance rates in \autoref{fig:draft_model_size_results}. These results provide more evidence for this finding: in all languages, increasing the draft model size improves acceptance rates. 
We see, however, that this does not necessarily translate to improved speed-ups.

\paragraph{Acceptance rates roughly correlate with resourcedness.} Following from the results in the prior section, we consistently see a positive correlation, where higher-resource languages tend to have higher acceptance rates. This provides evidence for Hypothesis 2, indicating that the curse of multilinguality is an important factor for speculative decoding. Of course, there are many other potential confounding factors, such as morphological complexity and relatedness to other languages.

Interestingly, there is a clear difference between the two tasks: translation has strong positive correlations for both models, while story generation shows somewhat weaker correlations. One explanation is that there is a small set of possible high-quality translations for a given source sentence, but a huge (approximately infinite) number of low-quality translations, and training the draft and verifier model on more data for a language makes both models converge towards that small high-quality set---and therefore the draft produces similar outputs to the verifier. Meanwhile, story generation is entirely open-ended with no ``correct'' answer, so the relationship between data and model divergence is weaker.



\subsection{Speculative Decoding Speed-Up Factors}
Next, we use acceptance rates from \S\ref{sec:acc_rates_ap} 
and  inference cost ratios from \autoref{fig:fwd_pass} to compute speed-up factors for each language and model pair, following the equation in \S\ref{sec:evaluation}. We report results in \autoref{fig:baseline_speedups} and make the following observations:

\paragraph{The Qwen model pair shows very small speed-ups.} For most low-resource languages, the Qwen model pair has a speed-up factor just slightly larger than 1. This result is not surprising, as these languages have very low acceptance rates, so the drafted tokens are rarely accepted by the verifier, and therefore speculative decoding would provide minimal improvements to overall generation speed. Meanwhile, the higher resource languages enjoy solid speed-ups for translation. Story generation has less of a clear trend, since acceptance rates seem to be a lot noisier.

\paragraph{Despite higher acceptance rates, Llama models have sub-baseline speed-up factors.} Though the prior section shows that Llama models have superior acceptance rates, the speed-up factors are consistently around or below 1, indicating that speculative decoding would actually cause a slow-down! This is due to an unfavorable inference cost ratio, as the 3b and 1b parameter Llama models are much closer in size. Furthermore, for the Qwen draft models of different sizes, we see a similar result: though acceptance rate unilaterally increases for larger draft models, speed-up factors do not consistently improve, or may even decrease due to an unfavorable cost ratio (\autoref{fig:draft_model_size_results}).

\begin{figure*}[tbh!]
    \centering
    \begin{subfigure}{0.49\textwidth}
        \includegraphics[width=1\linewidth]{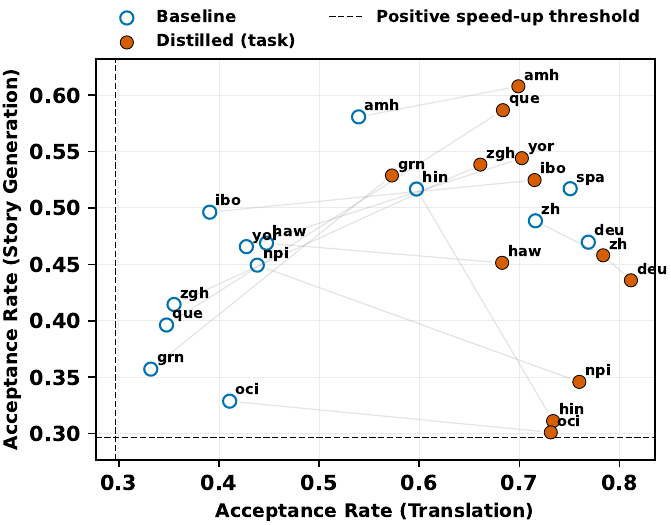}
    \end{subfigure}
    ~
    \begin{subfigure}{0.49\textwidth}
        \includegraphics[width=1\linewidth]{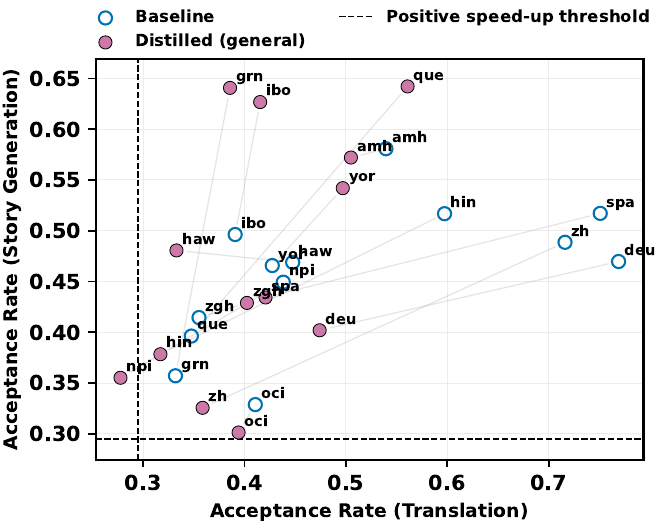}
    \end{subfigure}
    \caption{Acceptance rates for \textbf{Qwen-based models} on translation and story generation, before and after distillation on task-specific data (left) or general-domain data (right).}
    \label{fig:distillation_acceptance_rates}
\end{figure*}

\begin{figure*}[tbh!]
    \centering
    \begin{subfigure}{0.49\textwidth}
        \includegraphics[width=1\linewidth]{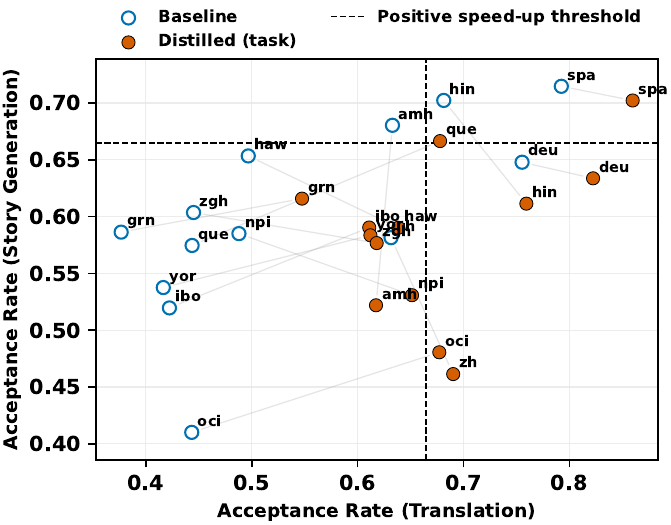}
    \end{subfigure}
    ~
    \begin{subfigure}{0.49\textwidth}
        \includegraphics[width=1\linewidth]{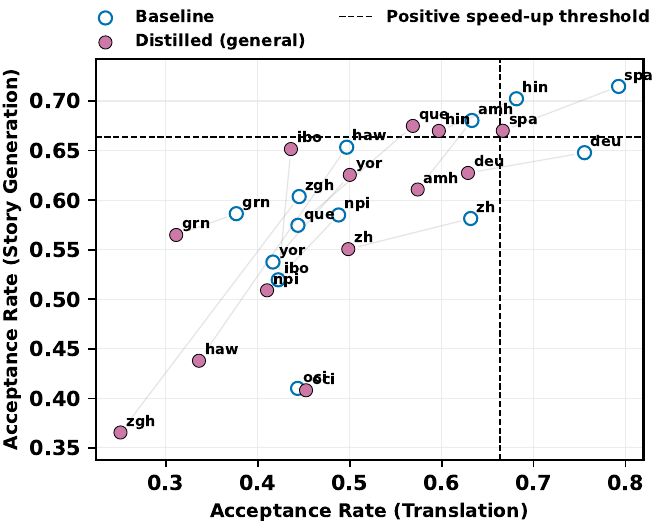}
    \end{subfigure}
    \caption{Acceptance rates for Llama-based models on translation and story generation, before and after distillation on task-specific data (left) or general-domain data (right).}
    \label{fig:llama_distillation_acceptance_rates}
\end{figure*}

These findings reinforce the critical tradeoff for speculative decoding: an ideal speed-up requires a draft model with a high acceptance rate and a favorable inference cost ratio. Thus, to improve speed-ups, we can either improve acceptance rates (\S\ref{sec:distillation}) or decrease draft model inference costs (\S\ref{sec:ngrams}).

\section{Distillation}
\label{sec:distillation}
\subsection{Methodology}

How can we reduce per-language divergence between draft and verifier? A common approach is 
\textbf{knowledge distillation} from  verifier to draft model \citep{hinton2015distillingknowledgeneuralnetwork}. We perform soft-target distillation (see \S \ref{sec:appendix_distillation}) from the verifier to the draft model minimizing the truncated cross-entropy loss.\footnote{We could optimize the LK divergence/acceptance rate directly, but in practice this is not effective \citep{goel2024directalignmentdraftmodel}.} 

An important methodological decision is the choice of dataset to distill on. We consider two typical approaches. First, we use the parallel corpora to construct translation prompts for on-policy distillation, where the teacher model generates a translation in the target language given a source sentence in English, representing a \textbf{task-specific distillation} approach. We use greedy decoding and only distill on the generated tokens (not the prompt). We hypothesize that this will reduce the divergence (and thus improve the acceptance rates) for other translation prompts, but will not generalize to other tasks in the same language.

Second, we use the monolingual corpora and perform off-policy distillation. Here, the teacher does not generate any outputs; rather, we simply compute the teacher logits over the existing text. We refer to this as the \textbf{general-domain distillation} approach. We hypothesize that while this setting may not improve acceptance rates as much as the previous setting, it should achieve a more balanced improvement across multiple tasks. 

We use the train splits from each dataset, and we select 5\% of these examples to use as a validation split. We perform soft-target distillation using the top 20 logits per-token, ignoring the tail of the logit distribution \citep{dasgupta2026dont}.

\begin{figure*}[tbh!]
    \centering
    \begin{subfigure}{0.49\textwidth}
        \includegraphics[width=1\linewidth]{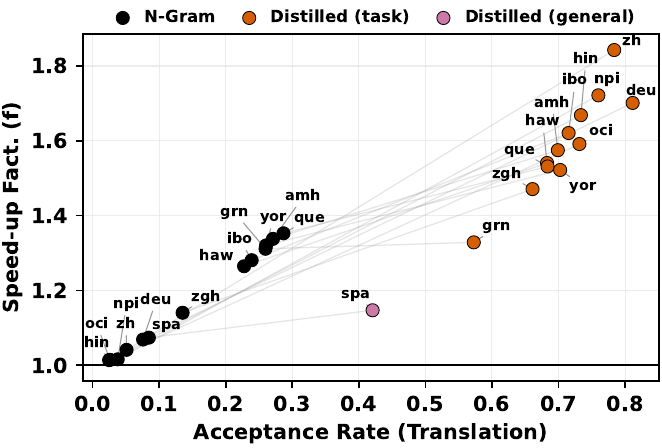}
    \end{subfigure}
    ~
    \begin{subfigure}{0.49\textwidth}
        \includegraphics[width=1\linewidth]{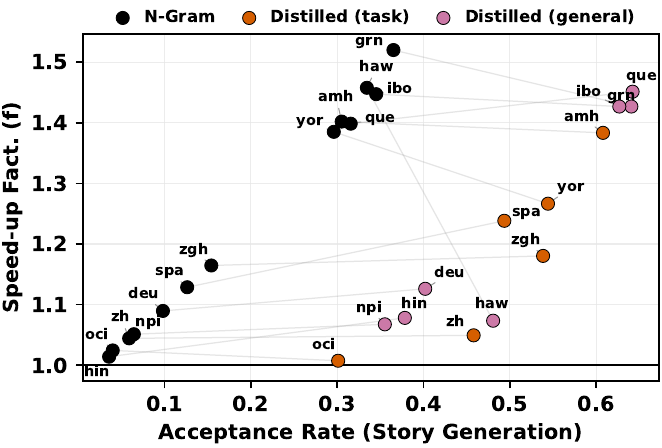}
    \end{subfigure}
    \caption{Acceptance rates (x-axis) and speed-up factors (y-axis) for both tasks using n-gram draft models, compared to the best \textbf{Qwen-based} distilled model from the prior section.}
    \label{fig:ngram_speedups}
\end{figure*}

\begin{figure*}[tbh!]
    \centering
    \begin{subfigure}{0.49\textwidth}
        \includegraphics[width=1\linewidth]{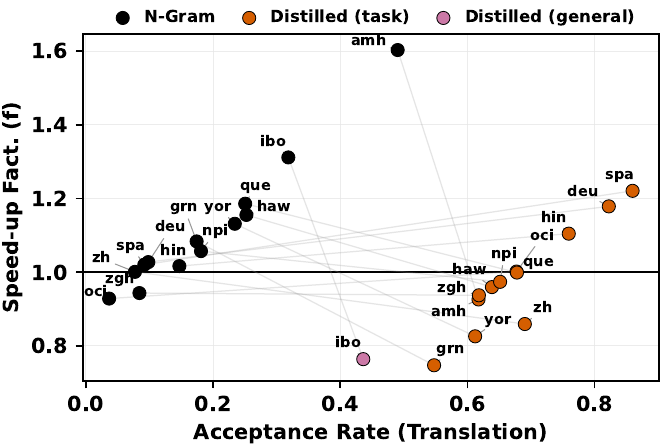}
    \end{subfigure}
    ~
    \begin{subfigure}{0.49\textwidth}
        \includegraphics[width=1\linewidth]{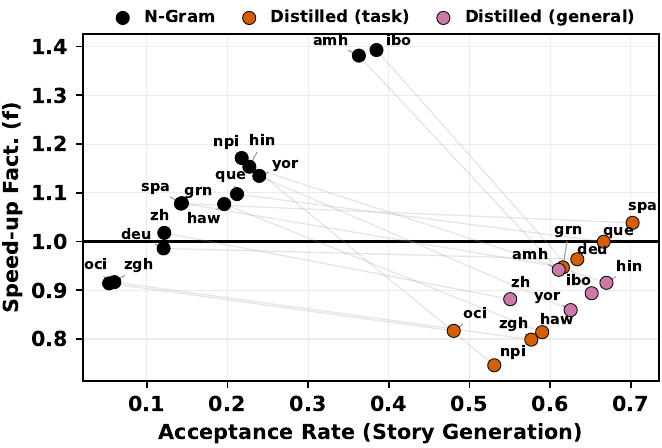}
    \end{subfigure}
    \caption{Acceptance rates (x-axis) and speed-up factors (y-axis) for both tasks using n-gram draft models, compared to the best \textbf{Llama-based} distilled model from the prior section.}
    \label{fig:llama_ngram_speedups}
\end{figure*}

\subsection{Results}
\autoref{fig:distillation_acceptance_rates} and \autoref{fig:llama_distillation_acceptance_rates} visualize the effect of distillation on acceptance rates for both tasks for Qwen and Llama, respectively. 
Since the draft model size does not change (only its parameters), increasing acceptance rates guarantees an increased speed-up factor, so we do not report speed-ups separately. We observe the following:

\paragraph{Task-specific distillation consistently improves acceptance rates on the same task, but generalizes poorly to a new task.} For every language, task-specific distillation improves acceptance rates for translation, in some cases by a massive amount (\texttt{npi}, \texttt{oci}). This result is intuitive, as distillation directly reduces the divergence between the draft and verifier model on in-distribution data. Furthermore, improving the KL divergence mathematically increases the lower bound on acceptance rates (\autoref{eq:bound}). We validate this bound for Qwen using the post-distillation KL divergence values and acceptance rates in \autoref{fig:pinsker}.

However, task-specific distillation does not generalize to the other task (story generation), and in fact worsens acceptance rates in many cases, supporting the results of \citet{sandler2025disparateimpactsspeculativedecoding}. This raises a key concern: unless there is a fully representative dataset of all the relevant tasks in a language, it is difficult to perform distillation without inadvertently worsening performance for some parts of the distribution. Thus, we turn to general-domain distillation as an alternate approach.

\paragraph{General-domain distillation has mixed effects on both tasks.} 
These results are noisier. Distillation improves acceptance rates on both tasks in some languages, but also may worsen performance on both. Naturally, the effectiveness of this form of distillation depends on the similarity between the monolingual data distribution and the output distribution for the chosen tasks, which likely varies between our corpora. Monolingual corpora for low-resource languages often represent few genres of text,
for example Bible translations or children's stories, and therefore these datasets may not provide uniform coverage over the full task distribution for a language \citep{marashian-etal-2025-priest}.

In summary, it is difficult to assemble a dataset for a low-resource language that provides ample coverage over the full task distribution. Furthermore, distillation on a limited dataset can cause increased divergence for other data---and thus is not always a reliable way to improve speculative decoding.




\section{N-Gram Draft Models}
\label{sec:ngrams}
\paragraph{Methodology.}
In the prior section, we demonstrated that improving acceptance rates for the full distribution of a low-resource language is generally infeasible. The other lever that can improve speed-up factors is inference cost ratio: a very cheap draft model may offer superior speed-ups, even with low acceptance rates. One such model used in prior work is a simple n-gram model, which predicts the next token given the prior $n-1$ tokens, using simple frequency statistics derived from a training corpus \citep{song-etal-2025-accelerated, chen-etal-2025-faster}. Naturally, n-gram models are far less effective at language modeling, as they have a small fixed-length context window; thus, we also expect to see a high divergence from a large transformer-based language model. However, these models are incredibly cheap to run (they can be implemented with a CPU-based hash table), typically orders of magnitude faster than a neural model as shown in \autoref{fig:fwd_pass}.

We train n-gram models on the monolingual corpora, using the Qwen
tokenizer to segment tokens. We test various $n$ values and find that bigrams (2-grams) produce the highest acceptance rates for all languages. During inference, we compute a simple logit distribution  for an $(n-1)$ length prefix from the conditional distribution observed during training. Our n-gram models are implemented with Python dictionaries, but could be optimized further.

\paragraph{Results.}
\autoref{fig:ngram_speedups} shows the acceptance rates and speed-up factors for n-gram models. We compare against the best distilled models from the prior section according to the speed-up factor. We see a clear trend:

\paragraph{N-gram models have far lower acceptance rates, but may have superior speed-up factors.} As predicted, n-gram models have low acceptance rates compared to neural draft models. However, due to substantial reduction in inference cost,
their speed-up factors are competitive with the distilled neural models. For Qwen, the n-gram models tend to be worse on translation but are often superior on story generation; for Llama the n-gram models are nearly always superior. The success on story generation indicates another benefit besides low inference cost: n-gram models are more robust to the data distribution,  generalizing to new tasks better than a neural model (trained on the same data). Therefore, depending on the cost ratio, acceptance rates, and task, n-gram models may be preferable for low-resource languages.

\section{Conclusion}
Our results confirm that speculative decoding may not be useful---and can even be detrimental---for low-resource languages, in part predicted by the curse of multilinguality. While distillation is an attractive way to reduce divergence, and thus improve acceptance rates, it is difficult to assemble a fully representative dataset, and there is often poor generalization to new tasks. Meanwhile, we demonstrate that simple n-gram models can provide superior speed-up factors in some cases thanks to the low cost of inference. In real-world usage, then, it would be beneficial to dynamically switch the draft model based on the language being generated, and in many cases an n-gram model may be the best choice. Alternately, more sophisticated speculative decoding techniques such as EAGLE \citep{pmlr-v235-li24bt} may be more effective for low-resource languages, but we predict that the divergence between the draft and verifier model would remain a limiting factor.

\section*{Limitations}
This work studies two families of models (Qwen 3.5 and Llama 3.2) and mainly focuses on a single verifier/drafter pair for each. Absolute results may vary across other size combinations and model families, but we expect the general trends to hold. Our work also studies a small set of thirteen languages, and results may differ for languages dissimilar to those thirteen. Finally, our work studies two realistic tasks in low-resource languages, and our results may not hold for tasks like mathematical reasoning (as in \citet{sandler2025disparateimpactsspeculativedecoding}).

\section*{Ethical Considerations}
This work suggests approaches to mitigate user experience disparities for disadvantaged languages. However, we recognize that work done without the involvement of language stakeholders risks poor alignment with the actual needs of speakers. In particular, we are aware that the generations in our evaluation suites are very poor quality, largely unusable for the desired tasks. Improving those generations is not a primary concern of this study, as our main focus is generation speed, but there remains a large gap in usability for many languages. Our work uses data sourced from existing open access datasets, and we defer to their ethical policies regarding data collection and processing. Finally, we recognize the severe environmental cost of training large-scale models, and have striven to use our resources efficiently. We used AI assistants for code review and to help modify plotting code.

\bibliography{custom,anthology-1,anthology-2}

\appendix


\begin{figure*}[tb!]
    \centering
    \begin{subfigure}{\linewidth}
    \includegraphics[width=1\linewidth]{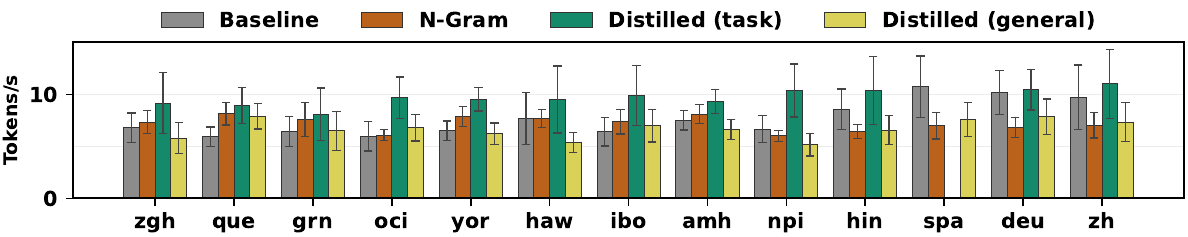}
    \end{subfigure}
    \begin{subfigure}{\linewidth}
    \includegraphics[width=1\linewidth]{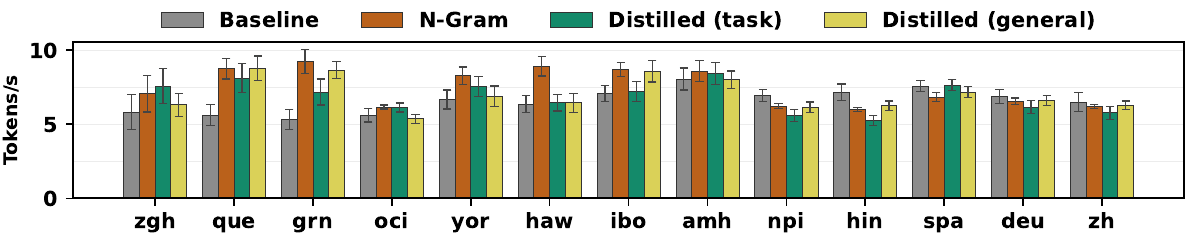}
    \end{subfigure}
    \caption{Tokens/second for translation (top) and story generation (bottom) prompts with Qwen models, over four experimental settings and thirteen languages. Error bars are standard deviation.}
    \label{fig:tps-transl}
\end{figure*}

\label{sec:chrf_acceptance} 
\begin{figure*}[tbh!]
    \centering
    \includegraphics[width=1\linewidth]{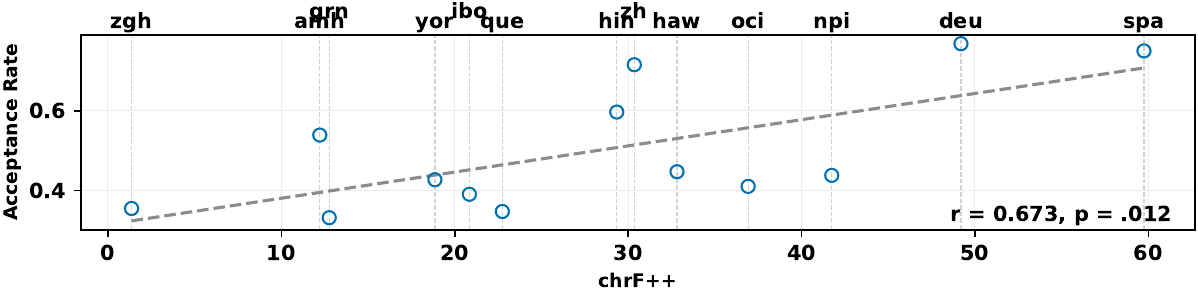}
    \caption{Relationship between the Qwen 9b model's translation quality (chrF) for each language and acceptance rates on translation prompts.}
    \label{fig:chrf_acceptance}
\end{figure*}

\section{Implementation and Resources}
We use the Hugging Face libraries for model implementations and tokenizers, Weights and Biases for experiment tracking, SacreBLEU for translation metrics, and NLTK to create topics for story generation. All experiments were run on a university cluster using H100 GPUs. In total, our experiments used roughly 500 GPU hours. 

We used LLM coding agents to assist in writing code, primarily for creating figures and reviewing human-written experimental code. We additionally used a chat-based LLM for finding literature and reviewing the final manuscript. However, we did not use LLMs to write any part of the paper. 

\section{Distillation Hyperparameters}
\label{sec:appendix_distillation}

All languages and models use the hyperparameters in \autoref{tab:distill_hparams} and bfloat16 precision, and we sweep the learning rate over $\{1e-4, 5e-5, 2e-5\}$ and select the best model by validation loss. For the general domain distillation approach, we truncate sequences to 128 tokens. For translation, we allow up to 128 generated tokens (excluding the prompt).

\begin{table}[htb!]
\small
    \centering
    \begin{tabular}{l|c}
    \toprule
        Total steps & 800 \\
        Batch size & 16  \\
        Grad. acc. steps & 4 \\
        Warm-up steps & 60 \\
        Optimizer & AdamW \\
        LR Schedule & cosine \\
        Weight decay & 0.01 \\
    \bottomrule
    \end{tabular}
    \caption{Distillation hyperparameters}
    \label{tab:distill_hparams}
\end{table}

\section{Dataset Sources}
\label{sec:sources}

We list the various data sources in \autoref{tab:sources}, as well as the per-language counts in \autoref{tab:mono_source_counts} and \autoref{tab:par_source_counts}.

\section{Additional Results}


\subsection{Throughput}

While speed-up factors provide the best estimate of performance, we also report throughput (\mbox{tokens/second}) metrics for the Qwen models, both tasks, and all four experimental settings in \autoref{fig:tps-transl}. We measure this as total generated tokens over total decode wall clock time, excluding the prefill:


\begin{equation}
\text{TPS} = \frac{\sum_{i=1}^{N} g_i}{\sum_{i=1}^{N} t_i}
\end{equation}

where $g_i$ is the number of output tokens for sentence $i$ and $t_i$ is its decode time. 


\subsection{Translation Quality}
\label{sec:appendix_translation_metrics}
We report chrF scores for the translations generated by the verifier model and the draft model before and after distilling on translation data in \autoref{tab:translation_scores}.

\begin{table}[h!]
    \small
    \centering
    \begin{tabular}{c | c c c | c c c}
        \toprule
        Lang & \multicolumn{3}{c|}{Qwen} & \multicolumn{3}{c}{Llama} \\
        & 9b & 0.8b & 0.8b dist & 3b & 1b & 1b dist \\
        \midrule
        amh & 13.3 & 0.9 & 11.3 & 5.6 & 1.3 & 5.4 \\
        grn & 12.6 & 9.4 & 12.4 & 10.1 & 8.0 & 9.7 \\
        haw & 28.9 & 9.8 & 26.8 & 15.2 & 8.6 & 15.2 \\
        ibo & 21.1 & 10.5 & 18.1 & 9.4 & 7.7 & 8.3 \\
        npi & 41.3 & 5.0 & 35.4 & 18.2 & 12.0 & 17.0 \\
        oci & 37.3 & 17.7 & 32.9 & 29.0 & 19.3 & 28.3 \\
        que & 22.0 & 13.1 & 22.0 & 13.0 & 11.2 & 12.2 \\
        yor & 18.3 & 8.7 & 16.6 & 9.7 & 7.2 & 7.8 \\
        zgh & 1.5 & 1.0 & 1.5 & 0.6 & 0.4 & 0.7 \\
        \hline
        deu & 49.9 & 37.8 & 42.7 & 44.5 & 35.6 & 41.5 \\
        spa & 58.7 & 44.4 & 51.5 & 54.8 & 42.9 & 52.1 \\
        hin & 28.7 & 16.8 & 26.7 & 24.5 & 19.1 & 23.1 \\
        zh & 31.4 & 19.9 & 25.6 & 20.7 & 13.5 & 18.8 \\
        \bottomrule
    \end{tabular}
    \caption{chrF scores for translations generated by the Qwen/Llama verifier models and draft models before and after distillation on translation data. Outputs are generated using sampling with the same parameters as the speculative decoding experiments.}
    \label{tab:translation_scores}
\end{table}

In \autoref{fig:chrf_acceptance}, we visualize the relationship between the Qwen 9b model's translation capability in a language, measured via the chrF score of generated translations, and acceptance rate. This provides another proxy for language resourcedness, though there are many other confounding factors, and we observe a strong correlation ($r=0.673$).


\subsection{Generated Story Examples}
We provide a few examples of generated stories in \autoref{fig:story-examples}.
\begin{figure}[h!]
\footnotesize
\setlength{\parskip}{0pt}

\noindent\hfill \textit{Topic: wooden cottage}
\vspace{2pt}\hrule\vspace{4pt}

\begin{tcolorbox}[
title=\textbf{Occitan},
colframe=BrickRed!80, colback=BrickRed!5,
fonttitle=\footnotesize\bfseries,
boxsep=2pt, left=4pt, right=4pt, top=2pt, bottom=2pt
]
La cabana de fusta josca las pichas,
Dormissiá sos pans de tèrra e las vègas.
Sa cabèssa èra plegada, coma una vella auca,
Los sos braçs de bala enredats dins lo vert.

L'escorècha, ligera coma l'aiga plegada,
Dançava amb la brisa qu'entrava pel pòrtal.
Cada chas de pluja l'auba mai fon,
Més sa màger de fusta ressonava l'
\end{tcolorbox}

\begin{tcolorbox}[
title=\textbf{Quechua},
colframe=NavyBlue!80, colback=NavyBlue!5,
fonttitle=\footnotesize\bfseries,
boxsep=2pt, left=4pt, right=4pt, top=2pt, bottom=2pt
]
Kanchisqan pisi llaktapi, wakin q'uñi qhapaq sunqupiqa, ch'askita yuyaywan huñusqa ruwayninpi uruwaq kaspa q'uñisqa. Chaymanta ch'ulluqta kaskaspa, qheya qusqa kaspa, ñuqu rakiy wasinqa uraypuni kaqta mirachispa kashan. T'aqa hawanpas, qullqinchikwan ch'ulluqninmi. Paypa sapann
\end{tcolorbox}

\vspace{8pt}
\noindent\hfill \textit{Topic: wild pet}
\vspace{2pt}\hrule\vspace{4pt}

\begin{tcolorbox}[
title=\textbf{Occitan},
colframe=BrickRed!80, colback=BrickRed!5,
fonttitle=\footnotesize\bfseries,
boxsep=2pt, left=4pt, right=4pt, top=2pt, bottom=2pt
]
Lo gat Negre de l'escòla de bòrde non es pas un domèstic, mas un viatge.
Un jorn de tèste, s'èra pichat dins lo jardin e non es pas tornat mai. La senhora Clara l'a cridat per l'amor, mas el fugia pels muros vèrs la montanha.
Avui, un jorn d'estiu, Clara l'a vist suls cims. Il èra amb un autre gat salvatge, de pel grisa e d'ull verds coma la fòrta.
\end{tcolorbox}

\begin{tcolorbox}[
title=\textbf{Quechua},
colframe=NavyBlue!80, colback=NavyBlue!5,
fonttitle=\footnotesize\bfseries,
boxsep=2pt, left=4pt, right=4pt, top=2pt, bottom=2pt
]
Ichaqa mankusqayki, qallariyta. K'anchanpi kachkan willka hurkuña, sarniwayniyuq ñawinchayninwan. Ima kutichaqtinqa k'anchayta qhawarichin, "Kimsa ñawikuna" nisqata. Llaman, hurkuñaqa manaraq ñawirinman rimayta atikunchu. Ch'iqun, q'umir raymi kaqlla pampallan. Willka hurkuñawan kuyurichis
\end{tcolorbox}



\caption{Samples of story generation for the topics \textit{wooden cottage} and \textit{wild pet}, sampled from the 9b Qwen 3.5 model.}
\label{fig:story-examples}
\end{figure}

\clearpage
\subsection{Distillation and Acceptance Rate}
\label{sec:kl_div}
When performing distillation on the same domain as the test data, we can compute a lower bound on the acceptance rate from the KL divergence between the teacher and student model. First, we use Theorem 3.5 from \citet{leviathan2023fastinferencetransformersspeculative} to get 
\begin{equation}
    \alpha = \sum_x \min(p(x), q(x))
\end{equation}
\noindent Let $A = \{x: p(x) \geq q(x)\}$ and $B = \{x: q(x) > p(x)\}$. Then, we get:
$$
    \alpha = \sum_{x\in A} q(x) + \sum_{x\in B} p(x)
$$
$$
    = \sum_{x\in A} p(x) - (p(x) - q(x)) + \sum_{x\in B} p(x)
$$
$$
    = \sum_{x\in A} p(x) + \sum_{x\in B} p(x) - \sum_{x \in A} (p(x) - q(x))
$$
\begin{equation}
    = 1 - \sum_{x \in A} (p(x) - q(x))
    \label{eq:acc}
\end{equation}
\noindent Next, we use the definition of \textit{total variation distance $D_{L,K}$}:
\begin{equation}
    D_{L,K}(P,Q) = \frac{1}{2}\sum_x \vert p(x)-q(x) \rvert
\end{equation}
\noindent Using the same definitions of $A$ and $B$,

\begin{equation}
\begin{split}
D_{L,K}(P,Q) = \frac{1}{2} \Bigl( &\sum_{x \in A} p(x) - q(x) \\
&+ \sum_{x \in B} q(x) - p(x) \Bigr)
\end{split}
\end{equation}

\noindent We show these two sums are equal:
$$
    \sum_{x \in A} p(x) - q(x) - \sum_{x \in B} q(x) - p(x)
$$
$$
= \sum_{x \in A} p(x) - q(x) + \sum_{x \in B} p(x) - q(x)
$$
$$
    = \sum_x p(x) - \sum_x q(x) = 0
$$
$$
    \therefore \sum_{x \in A} p(x) - q(x) = \sum_{x \in B} q(x) - p(x) = D_{L,K}(P,Q)
$$
\noindent Substituting into \autoref{eq:acc}, we get
\begin{equation}
    \alpha = 1 - D_{L,K}(P,Q)
\end{equation}
By Pinsker's inequality, we have an upper bound on the total variation distance:
\begin{equation}
    D_{L,K}(P,Q) \leq \sqrt {\frac{1}{2} D_{KL}(P \vert\vert Q)}
\end{equation}
\noindent where $D_{KL}$ is the KL-divergence. Finally, this gives the lower bound on acceptance rate:
\begin{equation}
    \alpha \geq 1-\sqrt {\frac{1}{2} D_{KL}(P \vert\vert Q)}
\end{equation}

\begin{figure}[tbh!]
    \centering
    \includegraphics[width=\linewidth]{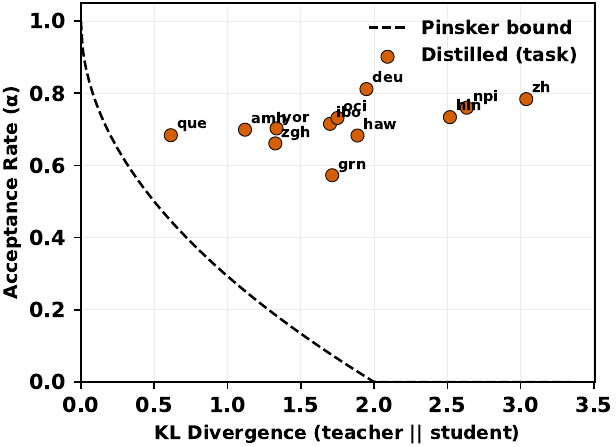}
    \caption{Relationship between KL divergence and acceptance rate for task-specific distilled Qwen models. The Pinsker bound shows the theoretical lower bound on acceptance rates given in \autoref{eq:bound}.}
    \label{fig:pinsker}
\end{figure}

We validate this bound against the task-specific distillation runs in \autoref{fig:pinsker}. We observe that all of the models well out-perform the bound, and there is not necessarily the negative correlation we would expect. However, this insight makes it possible to predict a lower bound on acceptance rate solely based on the distillation training run.

\begin{table*}[htb!]
\begin{threeparttable}
    \centering
    \begin{tabular}{l|l l c c}
        \toprule
        Code & Name & Source & Mono & Parallel  \\
        \midrule
        Aya & Aya Dataset & \citet{singh-etal-2024-aya} & \cmark \\
        FW2 & FineWeb2 & \citet{penedo2025fineweb2} & \cmark & \\
        Haw & Hawaiian Corpus Project & \citet{Doherty} & \cmark \\
        Lei & Leipzig Corpus Collections & \citet{goldhahn-etal-2012-building} & \cmark \\
        Men & MENYO-20k & \citet{adelani-etal-2021-effect} & & \cmark \\
        Ope & OpenCSG & \citet{yu2025opencsgchinesecorpusseries} & \cmark \\
        Opus & Opus & \citet{tiedemann-2012-parallel} & & \cmark \\
        O100 & OPUS-100 & \citet{zhang-etal-2020-improving} & & \cmark \\
        Que &  QuBERT & \citet{zevallos-etal-2022-introducing} & \cmark\\
        Tal & TALAM & \citet{IRCA} & \cmark \\
        Tat & Tatoeba & \citet{tiedemann-2020-tatoeba} & \cmark & \cmark  \\
        Tha & - & \citet{thapa-etal-2025-development} & \cmark &  \\
        Wub & Unified Amharic-English Corpus & GitHub\tnote{A} & & \cmark \\
        \bottomrule
    \end{tabular}
    \begin{tablenotes}
    \small
    \item[A] \url{https://github.com/wubet/unified-amharic-english-corpus}
    \end{tablenotes}
    \end{threeparttable}
    \caption{All of the sources used throughout our experiments. The \textit{Mono} and \textit{Parallel} columns indicate how we use the corpus; in some cases, a corpus has parallel sentences but we only use the text in the target language.}
    \label{tab:sources}
\end{table*}

\begin{table}[h!]
    \centering
        \begin{tabular}{l c l}
            \toprule
            \textbf{Language} & \textbf{Total Tokens} & \textbf{Sources} \\
            \midrule
            Tamazight [zgh] & 5.1M   & Tat, Tal \\
            Quechua [que]   & 642.0k & Que, Tat \\
            Guarani [grn]   & 398.5k & Lei \\
            Occitan [oci]   & 57.5k  & Lei \\
            Yoruba [yor]    & 449.6k & Aya \\
            Hawaiian [haw]  & 102.6k & Haw \\
            Igbo [ibo]      & 841.5k & Lei, Aya \\
            Amharic [amh]   & 148.5k & Aya \\
            Nepali [npi]    & 16.2M  & Tha, Tat, Aya \\
            \midrule
            Hindi [hin]     & 30.7M  & FW2, Aya \\
            Spanish [spa]   & 16.1M  & FW2, Aya \\
            German [deu]    & 15.2M  & FW2, Aya \\
            Chinese [zh]    & 17.6M  & Ope \\
            \bottomrule
        \end{tabular}
    \caption{Monolingual token counts for each language; see \autoref{tab:sources} for source abbreviations.}
    \label{tab:mono_source_counts}
\end{table}

\begin{table}[h!]
    \centering
        \begin{tabular}{l c l}
            \toprule
            \textbf{Language} & \textbf{Total Examples} & \textbf{Sources} \\
            \midrule
            Tamazight [zgh] & 5200 & Tat \\
            Quechua [que]   & 5200 & Opus, Tat \\
            Guarani [grn]   & 986  & Tat \\
            Occitan [oci]   & 4031 & Tat \\
            Yoruba [yor]    & 5200 & Men \\
            Hawaiian [haw]  & 121  & Tat \\
            Igbo [ibo]      & 1748 & Opus, Tat \\
            Amharic [amh]   & 5200 & Tat, Wub \\
            Nepali [npi]    & 3533 & Tat \\
            \midrule
            Hindi [hin]     & 5200 & O100 \\
            Spanish [spa]   & 5200 & O100 \\
            German [deu]    & 5200 & O100 \\
            Chinese [zh]    & 5200 & O100 \\
            \bottomrule
        \end{tabular}
    \caption{Parallel sentence counts for each language; see \autoref{tab:sources} for source abbreviations.}
    \label{tab:par_source_counts}
\end{table}

\end{document}